\documentclass{article}

\usepackage[preprint]{neurips_2025}

\usepackage[utf8]{inputenc} 
\usepackage[T1]{fontenc}    
\usepackage{hyperref}       
\usepackage{url}            
\usepackage{booktabs}       
\usepackage{amsfonts}       
\usepackage{nicefrac}       
\usepackage{microtype}      
\usepackage{times}
\usepackage{xspace}
\usepackage{amsmath}
\usepackage{booktabs}
\usepackage{enumitem}
\usepackage{multirow}
\usepackage{color,xcolor}
\usepackage{tabularx}
\usepackage{makecell}
\usepackage{lipsum}
\usepackage{geometry}
\usepackage{caption}
\usepackage{graphicx}
\usepackage{enumitem}  
\usepackage{listings}
\usepackage{pifont}
\usepackage{tcolorbox}
\usepackage{enumitem}
\usepackage{verbatim}
\usepackage{amsmath}
\usepackage{geometry}
\usepackage{booktabs} 
\definecolor{promptred}{RGB}{200,0,0}
\usepackage{bbding}
\setlist{nosep, after=\vspace{0pt}}

\title{Improving Medical Reasoning with Curriculum-Aware Reinforcement Learning}

\author{
Shaohao Rui$^{1,2,4}$, 
Kaitao Chen$^{3,4}$, 
Weijie Ma$^{2,3,4}$, 
Xiaosong Wang$^{2,4}$\thanks{Corresponding author: Xiaosong Wang. Email: \text{wangxiaosong@pjlab.org.cn}} \\
$^1$ Shanghai Jiao Tong University,
$^2$ Shanghai Innovation Institute \\
$^3$ Fudan University,
$^4$ Shanghai AI Laboratory \\
\url{https://github.com/shaohao011/MedCCO}
}

\begin{document}
\maketitle
\begin{abstract}
Recent advances in reinforcement learning with verifiable, rule-based rewards have greatly enhanced the reasoning capabilities and out-of-distribution generalization of VLMs/LLMs, obviating the need for manually crafted reasoning chains. Despite these promising developments in the general domain, their translation to medical imaging remains limited. Current medical reinforcement fine-tuning (RFT) methods predominantly focus on close-ended VQA, thereby restricting the model’s ability to engage in world knowledge retrieval and flexible task adaptation. More critically, these methods fall short of addressing the critical clinical demand for open-ended, reasoning-intensive decision-making. To bridge this gap, we introduce \textbf{MedCCO}, the first multimodal reinforcement learning framework tailored for medical VQA that unifies close-ended and open-ended data within a curriculum-driven RFT paradigm. Specifically, MedCCO is initially fine-tuned on a diverse set of close-ended medical VQA tasks to establish domain-grounded reasoning capabilities, and is then progressively adapted to open-ended tasks to foster deeper knowledge enhancement and clinical interpretability. We validate MedCCO across eight challenging medical VQA benchmarks, spanning both close-ended and open-ended settings. Experimental results show that MedCCO consistently enhances performance and generalization, achieving a 11.4\% accuracy gain across three in-domain tasks, and a 5.7\% improvement on five out-of-domain benchmarks. These findings highlight the promise of curriculum-guided RL in advancing robust, clinically-relevant reasoning in medical multimodal language models.
%

\end{abstract}

\section{Introduction}
\label{sec:intro}
Medical vision-language models (VLMs) have demonstrated significant advancements~\cite{moor2023med,wu2023towards,huang2025vision} in lesion detection and clinical diagnosis, driven primarily by supervised fine-tuning (SFT) on large-scale annotated datasets. Nonetheless, medical imaging tasks inherently demand more than mere accuracy in visual interpretation; they require transparent, clinically relevant rationales underpinning each diagnostic decision. Furthermore, the capacity for open-ended reasoning is frequently essential in real-world clinical settings, playing a critical role in trustworthy decision-making and timely therapeutic interventions.

Recent studies~\cite{deepseekai2025deepseekr1incentivizingreasoningcapability,qwq32b,du2025virgo,wang2025visualprm,huang2025vision} indicate that reinforcement learning (RL) is highly effective at enhancing sophisticated reasoning capabilities in large language models (LLMs) and general VLMs, obviating the necessity for meticulously crafted, high-quality long chain-of-thought (CoT~\cite{wei2022chain}) annotations. However, these promising results have yet to be thoroughly replicated or systematically validated within medical VLM contexts. Moreover, current RL-based methodologies applied to medical VLMs~\cite{lai2025med,pan2025medvlm} primarily target close-ended VQA tasks, often limited to perception-oriented assessments that probe basic visual comprehension~\cite{zuo2025medxpertqa}, e.g., modality and body parts. Such a restricted focus substantially constrains the models' ability to cultivate deeper, open-ended reasoning capabilities that are critical in clinical applications.

To tackle this issue, we propose \textbf{MedCCO}, the first multimodal reasoning framework that employs a curriculum-driven reinforcement learning paradigm for both medical close- and open-ended VQA tasks. Our objective is to leverage open-ended data within the reinforcement learning framework to enhance the model’s performance on close-ended reasoning, while simultaneously improving its capability in free-text reasoning to better address real-world clinical demands. We explore a couple of joint Group Relative Policy Optimization (GRPO) based RL training strategies and find that sequential knowledge injection through a curriculum-based approach yields superior results. Specifically, \textbf{MedCCO} is trained in a two-stage manner: it initially acquires domain-specific reasoning capabilities via close-ended medical VQA under reinforcement learning, and is subsequently adapted to more challenging open-ended tasks in a progressive fashion. This joint training scheme promotes both discriminative accuracy and generative flexibility, enabling the model to retrieve world knowledge and construct structured reasoning chains without relying on hand-crafted annotations. We conducted extensive experiments to assess the universality of \textbf{MedCCO} for medical VQA tasks, performing a systematic comparison across both close‑ended and open‑ended tasks. \textbf{MedCCO} consistently surpasses competitive baselines on in‑domain and out‑of‑domain test sets, and evaluations on the SLAKE~\cite{liu2021slake} benchmark further demonstrate its cross‑modal robustness.

Additionally, an ablation study on the joint RL paradigm revealed that a curriculum approach, first training on close‑ended examples, then progressively introducing open‑ended ones, outperforms training on both task types simultaneously. We attribute this gain to the conflicts between discrete and continuous reward signals and the differing difficulty levels of the two tasks. Together, these results suggest a promising reinforcement‑learning paradigm in which new reasoning capabilities can be injected to model sequentially, without sacrificing previously acquired performance.

To summarize, our contributions are three-fold:
\begin{enumerate}
  \item To the best of our knowledge, we propose the first multi-modal medical reasoning model capable of handling both close-ended and open-ended VQA tasks within a unified framework.
  \item We explore a curriculum reinforcement fine-tuning strategy that enables the VLM to learn from simple to complex tasks while retaining previously acquired knowledge.
  \item We conduct extensive experiments on both in-domain and out-of-domain medical VQA datasets. \textbf{MedCCO} achieves state-of-the-art performance and, on several benchmarks, matches or surpasses the performance of larger models significantly.
\end{enumerate}

\section{Related Work}
\label{sec:related_work}
\noindent{\textbf{General and Medical VLMs.}}  
Alignment-based methods such as CLIP~\cite{radford2021learning} and BLIP-2~\cite{li2023blip} paved the way for vision-language models. Subsequent work~\cite{liu2023visual, chen2024expanding, bai2025qwen2, li2023llava} demonstrates that instruction tuning with a few hundred thousand high-quality image–text pairs can unlock strong VQA and multimodal reasoning capabilities. This paradigm has been extended to the medical domain by models like Med-Flamingo~\cite{moor2023med}, LLaVA-Med~\cite{li2023llava_med}, and HuatuoGPT-Vision~\cite{chen2024huatuogpt}, which combine large-scale medical image–text alignment with domain-specific supervision to support accurate visual understanding and basic clinical diagnosis.

\noindent{\textbf{Reinforcement learning in LLMs/VLMs.}} Reinforcement learning has been widely adopted to align VLMs and LLMs with human preferences and mitigate hallucinations~\cite{sun2023aligning,yu2024rlhf,yu2024rlaif,zhao2023beyond,zhou2024aligning}. GRPO-based~\cite{deepseekai2025deepseekr1incentivizingreasoningcapability} methods have demonstrated strong effectiveness in enhancing the reasoning capabilities of VLMs/LLMs through rule-based rewards and got broadly verified in general visual reasoning scenarios~\cite{huang2025vision,liu2025seg,zhou2503r1,zhang2025r1,shen2025vlm}. However, such progress in the general domain has yet to be systematically validated in medical imaging. In this work, we extend this line of research to medical VQA by introducing a novel RL framework that uniquely integrates both close-ended and open-ended VQA data, and conduct comprehensive empirical evaluations to demonstrate its effectiveness.

\noindent{\textbf{Reinforcement Learning in Medical LLMs/VLMs.}} 
Recent studies have begun exploring reinforcement learning (RL) as a scalable alternative to supervised fine-tuning for enhancing medical reasoning in LLMs and VLMs. In LLMs, FineMedLM-o1~\cite{yu2025finemedlm} and MedReason~\cite{wu2025medreason} expand the training paradigm via test-time adaptation and structured knowledge supervision, while HuatuoGPT-o1~\cite{chen2024huatuogpt} leverages verifiable problems to improve reasoning through RL. In contrast, progress in VLMs remains limited. Recent efforts such as Med-R1~\cite{lai2025med} and MedVLM-R1~\cite{pan2025medvlm} show that rule-based RL can enhance generalization in medical VQA, but focus solely on close-ended questions, neglecting unified reasoning across modalities. To address this gap, our work applies GRPO-based RL within a unified framework encompassing both close-ended and open-ended VQA. To support this, we design task-specific rewards and adopt a curriculum training strategy that progressively enhances the model's reasoning capabilities.

\section{Methodology}
\label{sec:method}
\begin{figure}[ht]
    \centering
    \includegraphics[width=1\linewidth]{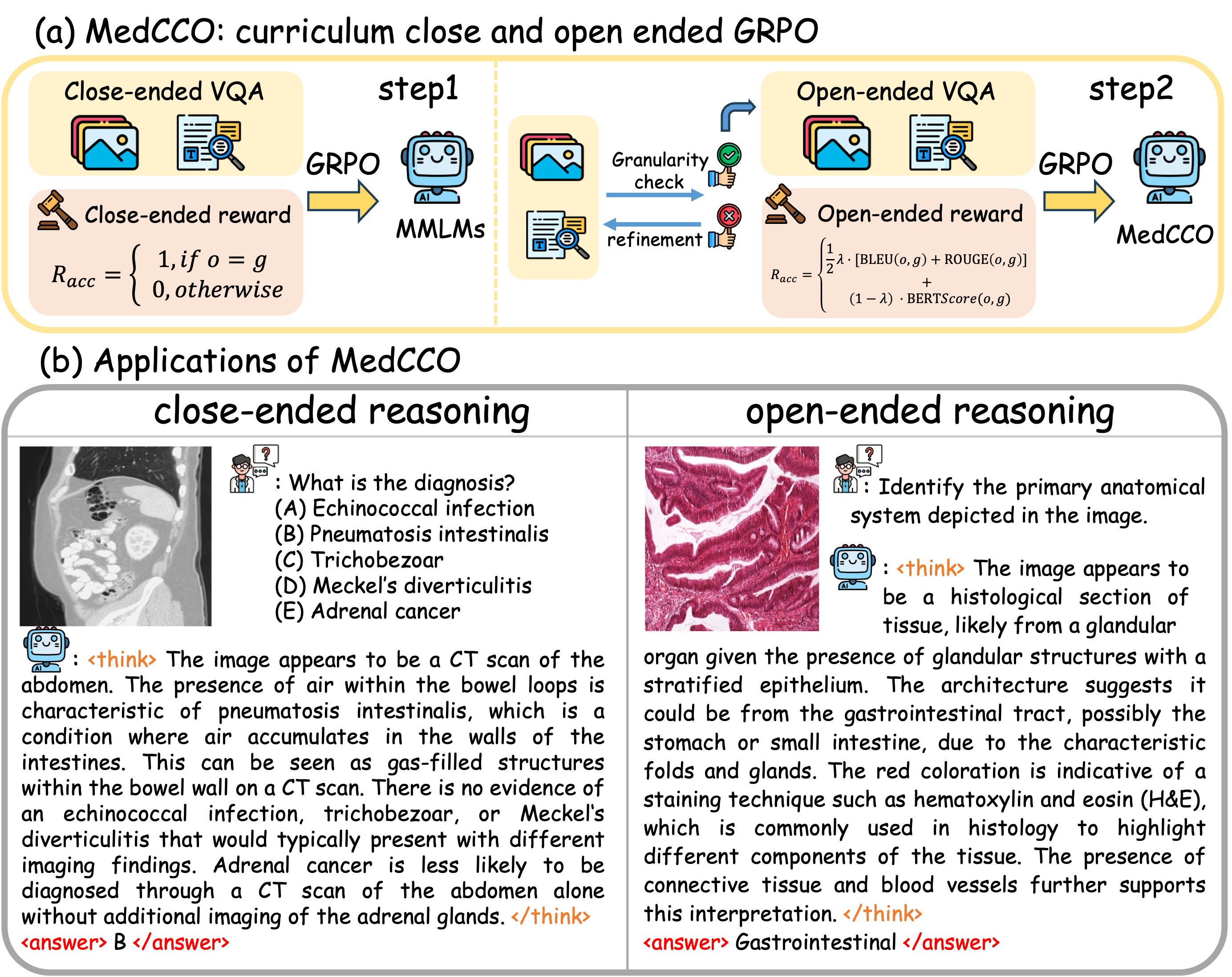}
    \caption{Overview of our MedCCO. MedCCO is first fine-tuned with GRPO on close-ended medical VQA to establish fundamental reasoning capabilities, then adapted to open-ended tasks to enable more advanced reasoning and knowledge retrieval. Prior to open-ended GRPO, we refine VQA data consistency to improve training efficiency.}
    \label{fig:overview}
\end{figure}

Medical VQA tasks can be broadly categorized into close-ended and open-ended formats. As shown in Figure~\ref{fig:overview} (b), close-ended VQA involves selecting from predefined options or verifying statements, while open-ended VQA requires free-form generation, demanding deeper reasoning over medical knowledge. In this work, we propose \textbf{MedCCO}, an extension of vanilla GRPO that integrates both types of VQA to enhance reasoning capabilities in medical VLMs. Following the curriculum learning strategy illustrated in Figure~\ref{fig:overview} (a), MedCCO is first trained on close-ended tasks with accuracy-based rewards, and subsequently fine-tuned on open-ended tasks, which are inherently more challenging. To stabilize reinforcement learning on open-ended data, we enhance VQA data consistency using an advanced VLM. By leveraging open-ended supervision, MedCCO improves not only close-ended performance but also fosters more autonomous and flexible reasoning.

\label{sec:preliminary}

\subsection{Backbone Learning Framework}
We build upon recent works~\cite{lai2025med,pan2025medvlm} that apply Group Relative Policy Optimization (GRPO) to improve the reasoning abilities of medical VLMs using close-ended VQA data. GRPO~\cite{shao2024deepseekmath} is a reinforcement learning algorithm similar to PPO~\cite{schulman2017proximal}, with two key differences: (1) GRPO operates in a value-free regime by computing generalized advantage estimation (GAE) using group-relative rewards; and (2) it employs verifiable rule-based outcomes as rewards instead of relying on pre-trained reward models. At each training step, the model generates $G$ candidate responses $\{o_i\}_{i=1}^{G}$ from the current policy $\pi_{\theta_{\text{old}}}$. Each output is assigned a scalar reward $r_i$ based on rule-based criteria. The advantage is normalized by the mean and standard deviation of all group rewards:

\begin{align}
\label{eq:grpo}
A_i = \frac{r_i - \text{mean}(\{r_j\}_{j=1}^G)}{\text{std}(\{r_j\}_{j=1}^G)},
\end{align}

The GRPO objective is defined as:
{\small
\begin{align}
\mathcal{L}_{\mathrm{GRPO}}(\theta) = 
& -\frac{1}{\sum_{i=1}^{G} |o_i|} \sum_{i=1}^{G} \sum_{t=1}^{|o_i|} \Bigg[
\min\left( 
\frac{\pi_\theta(o_{i,t} \mid q, o_{i,<t})}{\pi_{\theta_{\text{old}}}(o_{i,t} \mid q, o_{i,<t})}, \right. \nonumber \\
& \left. \text{clip}\left( 
\frac{\pi_\theta(o_{i,t} \mid q, o_{i,<t})}{\pi_{\theta_{\text{old}}}(o_{i,t} \mid q, o_{i,<t})}, 
1 - \epsilon, 
1 + \epsilon 
\right) 
\right) \hat{A}_{i,t} - \beta \mathbb{D}_{\mathrm{KL}}[\pi_\theta \| \pi_{\text{ref}}] \Bigg],
\end{align}
}

where $\mathbb{D}_{\text{KL}}(\pi_\theta \| \pi_{\text{ref}})$ serves as a regularization term to penalize divergence from the reference policy $\pi_{\text{ref}}$, with $\beta$ controlling its strength.

\subsection{Multi-reward Policy}

\begin{table}[t]
\centering
\begin{tabular}{p{0.95\linewidth}}
\toprule
You are a helpful assistant. \{Question\} Output the thinking process in <think> </think> and final answer in <answer> </answer> tags. The output answer format should be as follows: <think> reasoning process here </think><answer> answer here (Do not provide any explanation) </answer> Please strictly follow the format. \\
\bottomrule
\end{tabular}
\caption{Prompt template used in MedCCO. The placeholder \texttt{\{Question\}} is replaced with specific reasoning questions during training.}
\label{tab:prompt-template}
\end{table}

To effectively guide reinforcement learning and prevent reward hacking, we design a multi-dimensional reward schema targeting three aspects: correctness, semantic alignment, and format adherence. Each component targets a distinct behavioral dimension. During training, models are prompted using the instruction shown in Table~\ref{tab:prompt-template}, and structured responses are parsed to compute corresponding rewards.

\noindent\textbf{Close-ended Reward.} For close-ended tasks, we use a binary reward to enforce strict correctness:
\begin{align}
R_{\text{close}}(o, g) =
\begin{cases}
1, & \text{if } o = g, \\
0, & \text{otherwise},
\end{cases}
\end{align}
where \( o \) is the predicted answer and \( g \) is the ground truth.

\vspace{3pt}
\noindent\textbf{Open-ended Rewards.} For open-ended responses, we design a hybrid reward function that jointly captures surface-level fidelity and deep semantic alignment. Given that the ground-truth answers are relatively short (with a mean length of approximately 5 tokens), we employ BLEU-1 and ROUGE-1 to assess lexical overlap, while BERTScore is used to evaluate semantic similarity.
\begin{align}
R_{\text{open}}(o, g) = \frac{1}{2} \lambda \cdot (\text{BLEU}_1(o, g) + \text{ROUGE}_1(o, g)) + (1 - \lambda) \cdot \text{BERTScore}(o, g),
\end{align}
where \( \lambda \in [0, 1] \) controls the trade-off between lexical and semantic metrics.

\vspace{3pt}
\noindent\textbf{Format Reward.} To ensure structured output, we impose a format reward (denoted as $R_{\text{format}}$)that checks for compliance with required tags. Specifically, the reasoning content must be enclosed in \texttt{<think>}~···~\texttt{</think>} and the answer in \texttt{<answer>}~···~\texttt{</answer>} tags respectively.

\noindent{\textbf{Total Reward.}} For each sample, the total reward is computed as $\gamma R + (1-\gamma) R_{\text{format}}$, where $R$ denotes $R_{\text{close}}$ for close-ended and $R_{\text{open}}$ for open-ended questions.

\subsection{Joint Reinforcement Learning Training Strategies}
To combine close-ended and open-ended VQA tasks under a unified reinforcement learning framework, we explore two gradient-based reinforcement policy optimization strategies based on GRPO.

\noindent{\textbf{{Direct joint GRPO via gradient re-weighting.}} We first investigate joint GRPO by simultaneously optimizing over both task types. Due to the inherently different reward structures: discrete for close-ended and continuous for open-ended, the resulting reward advantages exhibit distinct variance characteristics, which in turn affect the gradient magnitudes during optimization. To address this, we re-weight the gradients of each task \emph{within the same mini-batch} to balance their contributions. Specifically, given a mini-batch \(\mathcal{B} = \mathcal{B}_{\mathrm{c}} \cup \mathcal{B}_{\mathrm{o}}\), where \(\mathcal{B}_{\mathrm{c}}\) and \(\mathcal{B}_{\mathrm{o}}\) denote close- and open-ended samples respectively, we compute per-sample gradients \( g^{(k)} = \nabla_\theta{\mathcal{L}_{\rm GRPO}(\theta)} \), and task-wise averaged gradients are:
\begin{align}
\bar g_{\mathrm{c}} &= \frac{1}{|\mathcal{B}_{\mathrm{c}}|} \sum_{i\in\mathcal{B}_{\mathrm{c}}} g^{(i)}, \quad
\bar g_{\mathrm{o}} = \frac{1}{|\mathcal{B}_{\mathrm{o}}|} \sum_{j\in\mathcal{B}_{\mathrm{o}}} g^{(j)},
\end{align}
and the final combined gradient is:
\begin{equation}
\label{eq:combo}
\mathcal{G} = \alpha \cdot \bar g_{\mathrm{c}} + (1 - \alpha) \cdot \bar g_{\mathrm{o}}, \quad \alpha = \frac{|\mathcal{B}_{\mathrm{o}}|}{|\mathcal{B}_{\mathrm{c}}| + |\mathcal{B}_{\mathrm{o}}|}
\end{equation}
where the mixing coefficient \( \alpha \in (0, 1) \) is statically set based on the relative batch sizes of close-ended and open-ended samples, ensuring proportionally balanced gradient contributions across tasks.

\noindent{\textbf{Curriculum GRPO.}}  
We further explore a curriculum-based strategy to address the challenges of applying RL to open-ended VQA. Unlike close-ended tasks with well-defined answer spaces, open-ended VQA requires the VLM to generate free-form responses that align closely with ground truth—an inherently difficult task in the medical domain where domain knowledge is essential. This often results in unstable and inefficient learning. Inspired by curriculum learning~\cite{bengio2009curriculum}, we adopt a progressive training scheme to gradually build reasoning capability. As illustrated in Figure~\ref{fig:overview}, the model is first trained on close-ended questions using GRPO to establish a stable policy, followed by fine-tuning on open-ended data. Empirically, this curriculum-driven GRPO approach stabilizes learning and consistently improves reasoning performance across both question types.

\subsection{VQA Data Quality Refinement}
Prior to using open-ended VQA data for GRPO, as depicted in Figure~\ref{fig:overview} (a), we perform a VQA consistency check and refinement. A major issue observed in current open-ended medical VQA datasets is the inconsistency in granularity between questions and answers, which impairs the learning process. While supervised fine-tuning primarily maps input-output patterns, reinforcement learning requires the model to align its reasoning with semantic details. For instance, in SLAKE~\cite{liu2021slake} dataset, the question \textit{“How was the image taken?”} may yield answers like \textit{“CT”} or \textit{“axial”}, which differ in specificity and introduce ambiguity.

To address this, we implement a VQA-Consistency Auditor (i.e. Qwen2.5-VL~\cite{bai2025qwen2}-72B) that refines noisy VQA pairs based on three core principles: (1) \textbf{Consistency evaluation:} ensuring the question comprehensively captures the semantic content of the answer; (2) \textbf{Open-ended enforcement:} maintaining free-form phrasing to support descriptive outputs; and (3) \textbf{Granularity matching:} aligning the specificity of the question with the answer to reduce under- or over-generalization. Detailed criteria and prompts for this refinement process are included in the supplementary materials.

\section{Experiments}
\label{sec:exp}
\noindent\textbf{Training Datasets.}
We utilize three publicly available medical VQA datasets—VQA-RAD~\cite{lau2018dataset}, SLAKE~\cite{liu2021slake} (English part), and PathVQA~\cite{he2020pathvqa} as in-domain training datasets. Together, these datasets provide a total of 14,379 close-ended and 12,996 open-ended question-answer pairs, comprising approximately 27k training samples. For in-domain evaluation, we adhere to the official data splits provided by each dataset. To evaluate cross-modal generalization, we leverage the SLAKE dataset~\cite{liu2021slake}, which spans three imaging modalities: X-ray, MRI, and CT, each comprising both open-ended and close-ended samples. We train the model on a single modality and evaluate it across others' test sets. More details are in the supplementary materials.

\noindent
\textbf{Benchmarks.} We adopt the same experimental settings and benchmarks as in \cite{chen2024huatuogpt}, with additional focus on recent reasoning datasets (enhanced coverage and difficulty), i.e., test sets of VQA-RAD~\cite{lau2018dataset}, SLAKE~\cite{liu2021slake}, PathVQA~\cite{he2020pathvqa}, PMC-VQA~\cite{zhang2023pmc}, and the Health \& Medicine track of MMMU~\cite{yue2024mmmu} for multi-modality handling.
In addition, we include Quilt-VQA~\cite{seyfioglu2024quilt} for out-of-domain evaluation on open-ended questions, and MedXpertQA~\cite{zuo2025medxpertqa}(MM part), a recently proposed benchmark emphasizing complex medical reasoning.


\noindent{\textbf{Baselines.}} We compare our MedCCO's performance with Gerneral VLMs: Yi-VL~\cite{young2024yi}, LLaVA-v1.6~\cite{liu2024llavanext} and Qwen2.5-VL~\cite{bai2025qwen2}; Medical VLMs: Med-Flamingo~\cite{moor2023med}, RadFM~\cite{wu2023towards}, LLaVA-Med~\cite{li2023llava_med}, and HuatuoGPT-Vision~\cite{chen2024huatuogpt}. 
There is limited research on applying reinforcement learning to enhance reasoning in medical VLMs. Recent efforts such as MedR1~\cite{lai2025med} and MedVLM-R1~\cite{pan2025medvlm} employed vanilla GRPO for different medical tasks. Thus, we also compare our approach with vanilla GRPO fine-tuning, denoted as Qwen2.5-VL (GRPO), trained on close-ended and refined open-ended data.

\noindent{\textbf{Implementation Details.}}
Training is conducted using 4$\times$H100 GPUs (80GB VRAM each) with PyTorch, leveraging FlashAttention-2 for computational efficiency. GRPO-related experiments are implemented using the \texttt{verl}~\cite{sheng2024hybridflow} framework to accelerate the training. We adopt the Qwen2.5-VL~\cite{bai2025qwen2}-7B-Instruct model and its 3B variant as the backbone models. The training uses a total batch size of 64, with a learning rate of $1\text{e}^{-6}$. The KL penalty coefficient is set to $\beta = 0.01$, with $\lambda = 0.7$ and $\gamma = 0.8$. For inference during GRPO, we deploy the model using \texttt{vllm}~\cite{kwon2023efficient} on 2 GPUs, generating 10 completions per sample, corresponding to the group size $G$ in GRPO. The maximum number of input pixels is capped at 401,408, and the temperature for \texttt{vllm} sampling is set to 1.0. GRPO training is conducted for 1 epoch and about 8 hours. For supervised fine-tuning (SFT), all experiments are performed using \texttt{LLaMAFactory}~\cite{zheng2024llamafactory}, with its default configuration. More details can be found in the supplemental material.

\subsection{Comparison Results}
\begin{table*}[t]
  \small
  \centering
  \setlength{\tabcolsep}{2.5pt}
  \resizebox{1\linewidth}{!}{
  \begin{tabularx}{\textwidth}{%
      l
      |cc
      cc
      cc
      c|
      cc
      c
      cc
      c
    }
    \toprule
    \multirow{3}{*}{Model}
      & \multicolumn{7}{c|}{In-domain test}
      & \multicolumn{6}{c}{Out-of-domain test} \\
    \cmidrule(lr){2-8} \cmidrule(lr){9-14}
      & \multicolumn{2}{c}{VQA-RAD}
      & \multicolumn{2}{c}{SLAKE}
      & \multicolumn{2}{c}{Path-VQA}
      & Avg.
      & \multicolumn{2}{c}{Quilt-VQA}
      & PMC
      & \multicolumn{2}{c}{MedXpert}
      & Avg \\
      & c. & o.   & c. & o.   & c. & o.   &      & c. & o.   & c.  & R.c. & U.c. &      \\
    \midrule
    \multicolumn{14}{c}{\itshape\textbf{General VLM}} \\[3pt]
    Yi-VL~\cite{young2024yi}-34B            & 53.0 & 22.4 & 58.9 & 33.9 & 47.3 & 12.9 & 38.1 & 56.0 & 13.2 & 39.5 & 19.9 & 20.7 & 29.9 \\
    LLaVA-v1.6~\cite{liu2024llavanext}-7B        & 52.6 & 19.8 & 57.9 & 37.6 & 47.9 & 12.6 & 38.1 & 58.3 &  8.7 & 35.5 & 20.7 & 20.6 & 28.8 \\
    LLaVA-v1.6~\cite{liu2024llavanext}-13B       & 55.8 & 24.0 & 58.9 & 44.5 & 51.9 & 12.8 & 41.3 & 57.4 & 24.5 & 36.6 & 19.5 & 18.1 & 31.2 \\
    LLaVA-v1.6~\cite{liu2024llavanext}-34B       & 58.6 & 24.1 & 67.3 & 44.6 & 59.1 & 15.0 & 44.8 & 62.4 & 23.7 & 44.4 & 20.6 & \textcolor{red}{25.5} & 35.3 \\
    Qwen2.5-VL~\cite{bai2025qwen2}-7B        & 67.3 & 32.2 & 71.6 & 40.2 & 65.5 & 17.2 & 49.0 & 54.8 & 29.0 & 50.4 & 20.6 & 23.1 & 35.6 \\
    \midrule
    \multicolumn{14}{c}{\itshape\textbf{Medical VLM}} \\[3pt]
    Med-Flamingo~\cite{moor2023med}-7B      & 45.4 & 29.3 & 43.5 & 30.1 & 54.7 & 28.7 & 38.6 & 62.1 & 22.3 & 23.3 & 19.0 & 20.0 & 29.3 \\
    RadFM~\cite{wu2023towards}-13B            & 50.6 & 34.0 & 34.6 & 44.2 & 38.7 & 19.9 & 37.0 & 60.7 & 21.5 & 25.9 & 19.8 & 19.6 & 29.5 \\
    LLaVA-Med~\cite{li2023llava_med}-7B         & 51.4 & 10.1 & 48.6 &  6.6 & 56.8 &  8.4 & 30.3 & 63.0 & 29.3 & 24.7 & 20.5 & 19.5 & 31.4 \\
    HuatuoGPT-Vision~\cite{chen2024huatuogpt}-8B  & 63.8 & \textcolor{blue}{36.0} & 74.5 & 47.0 & 59.9 & 23.2 & 50.7 & \textcolor{blue}{63.9} & \textcolor{blue}{38.5} & \textcolor{blue}{52.7} & 20.4 & 22.9 & \textcolor{blue}{39.6} \\
    \midrule
    \multicolumn{14}{c}{\itshape\textbf{Finetuned VLM}} \\[3pt]
    Qwen2.5-VL~\cite{bai2025qwen2}-7B (SFT)  & \textcolor{blue}{71.3}& 27.8 & 78.6 & \textcolor{blue}{50.8} & \textcolor{red}{87.8} & \textcolor{red}{33.6} & \textcolor{blue}{58.3} & 60.9 &  8.9 & 49.2 & 20.2 & 20.4 & 31.9 \\
    Qwen2.5-VL~\cite{bai2025qwen2}-7B (GRPO)& 70.5 & 29.8 &\textcolor{blue}{79.3} & 40.2 & \textcolor{blue}{82.8} & 27.8 & 55.1 & 50.2 & 28.4 & 51.2 & \textcolor{blue}{21.2} & 21.7 & 34.5 \\
    MedCCO-7B           & \textcolor{red}{76.3} & \textcolor{red}{40.0} & \textcolor{red}{79.4} & \textcolor{red}{65.7} & \textcolor{blue}{82.8} & \textcolor{blue}{29.6} & \textcolor{red}{62.3} & \textcolor{red}{69.4} & \textcolor{red}{39.3} & \textcolor{red}{53.2} & \textcolor{red}{23.2} & \textcolor{blue}{23.6} & \textcolor{red}{41.7} \\
    \bottomrule
  \end{tabularx}}
    \caption{Performance of our MedCCO, Medical VLMs and General VLMs on three in-domain datasets and three out-of-domain datasets. The best and second-best results in each column are highlighted in \textcolor{red}{red} and \textcolor{blue}{blue}, respectively. c.: close-end accuracy. o.: open-ended metrics (combined BLEU1, ROUGE1 and BERTScore.). R: reasoning, U: understanding.}
  \label{tab:medical_vqa_results}
\end{table*}

As shown in Table~\ref{tab:medical_vqa_results}, MedCCO sets a new state-of-the-art across a wide range of medical VQA benchmarks, including both in-domain and out-of-domain evaluations. It achieves an average accuracy of 62.3\% on in-domain datasets and 41.7\% on out-of-domain ones, outperforming existing generalist and domain-specific baselines on 8 out of 11 sub-tasks. Compared to the second-best model, HuatuoGPT-Vision~\cite{chen2024huatuogpt}-8B, MedCCO achieves a 11.6\% gain in in-domain accuracy while maintaining stronger generalization to novel tasks such as MedXpertQA~\cite{zuo2025medxpertqa} and Quilt-VQA~\cite{seyfioglu2024quilt}. These findings validate the effectiveness of our unified curriculum-guided GRPO approach in enhancing medical reasoning across settings.


As shown in Table~\ref{tab:mmmu_results}, MedCCO achieves the highest overall accuracy of 59.3\% on the MMMU benchmark, surpassing all general-purpose, medical-specific, and fine-tuned VLMs. Notably, it outperforms HuatuoGPT-Vision~\cite{huang2025vision}-8B by over 10.2\% in absolute accuracy, despite the latter being trained on more than 1M image–text pairs. These results underscore MedCCO’s strength in comprehensive multimodal understanding and its capacity for integrating visual and medical knowledge effectively.

\begin{figure}[t]
    \centering
    \includegraphics[width=0.9\linewidth]{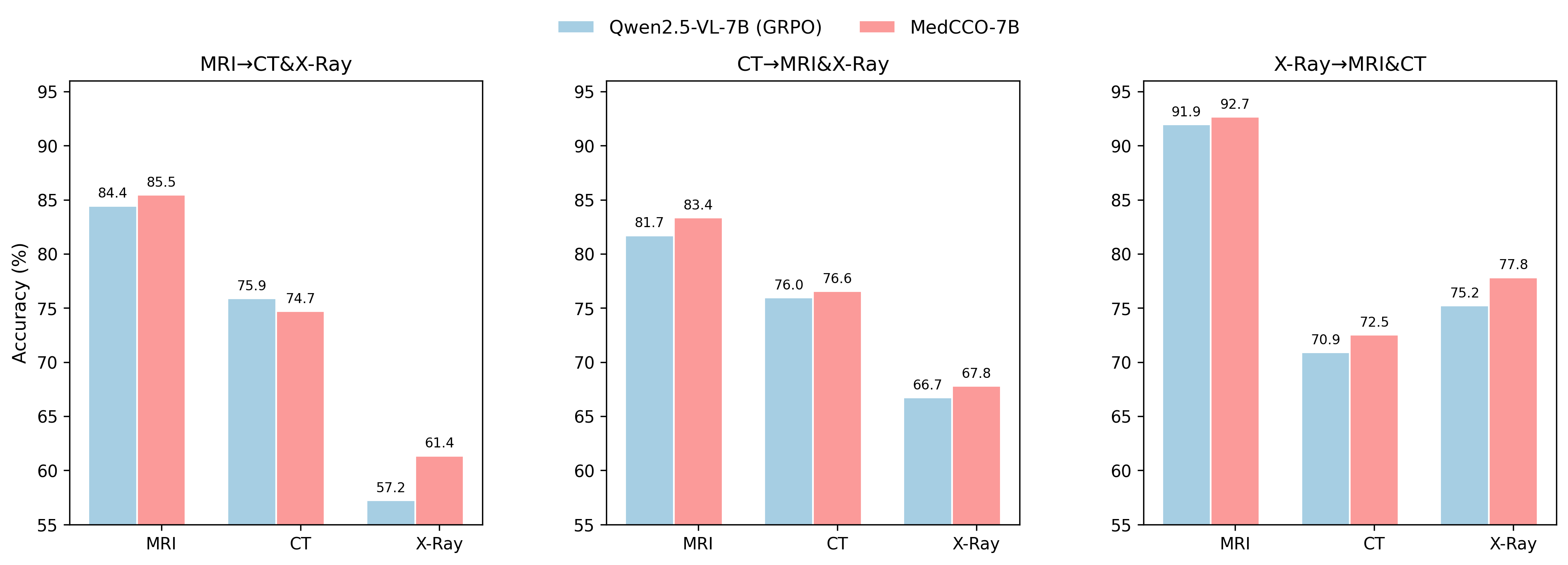}
    \caption{Cross-modal performance on SLAKE~\cite{liu2021slake}, with each model trained on a single modality and evaluated across all modalities for in- and cross-modal comparison.}
    \label{fig:slake_crossmodal}
\end{figure}
To assess cross-modal transferability, we conduct experiments on the SLAKE~\cite{liu2021slake} dataset by training on a single modality and evaluating on all three (MRI, CT, X-ray). Test sets refer to the union of the validation and test splits from the official SLAKE~\cite{liu2021slake} dataset, which differs from the configuration used in Table~\ref{tab:medical_vqa_results}. As shown in Figure~\ref{fig:slake_crossmodal}, MedCCO-7B consistently outperforms Qwen2.5-VL~\cite{bai2025qwen2}-7B using GRPO at all transfer scenarios. Most notably, under the most challenging X-ray $\rightarrow$ MRI \& CT setting, MedCCO attains an accuracy of 92.7\% on MRI and 72.5\% on CT, achieving the best results across all modalities.

\noindent{\textbf{Qualitative Analysis.}} Figure~\ref{fig:case_study} showcases representative examples of MedCCO’s reasoning capability. In the open-ended cases (a) and (b), the model demonstrates accurate anatomical identification and physiological interpretation. In (a), it recognizes the lungs in a chest X-ray and correctly infers their primary function as breathe. In (b), the model identifies two kidneys in a cross-sectional CT scan based on their spatial location and morphological features. For the close-ended example (c), MedCCO handles a clinically grounded diagnostic task by selecting surgical excision as the next management step for a shoulder mass, justifying its decision with multi-modal evidence, including MRI findings and absence of metastasis on CT. These examples highlight the model’s ability to integrate visual cues with medical knowledge to generate both descriptive and decision-oriented responses. For additional qualitative examples, please refer to the supplementary materials.

\begin{table*}[t]
\centering
\resizebox{0.8\linewidth}{!}{
\begin{tabular}{l|ccccc|c}
\toprule
\textbf{Model} & \makecell{BMS} & \makecell{CM} & \makecell{DLM} & \makecell{P} & \makecell{PH} & \makecell{MMMU \\ Health \& Medicine} \\
\midrule
\multicolumn{7}{c}{\textit{General VLM}} \\

Yi-VL~\cite{young2024yi}-34B & 49.4 & 48.9 & 43.2 & 40.5 & 32.0 & 41.5 \\
LLaVA-v1.6~\cite{liu2024llavanext}-7B & 40.5 & 36.9 & 32.1 & 32.3 & 26.9 & 33.1 \\
LLaVA-v1.6~\cite{liu2024llavanext}-13B & 53.6 & 46.7 & 33.3 & 22.2 & 40.0 & 39.3 \\
LLaVA-v1.6~\cite{liu2024llavanext}-34B & 56.4 & 56.0 & \textcolor{blue}{46.9} & 46.7 & 41.7 & 48.8 \\
LLaVA-v1.5-LLaMA3-8B & 42.3 & 44.0 & 37.0 & 34.7 & 35.2 & 38.2 \\
Qwen2.5-VL~\cite{bai2025qwen2}-7B & 50.0 & \textcolor{blue}{63.3} & 33.3 & 59.3 & 53.3 & 51.7 \\
\midrule
\multicolumn{7}{c}{\textit{Medical VLM}} \\
Med-Flamingo~\cite{moor2023med} & 29.6 & 28.1 & 24.8 & 25.3 & 31.2 & 28.3 \\
RadFM~\cite{wu2023towards} & 27.5 & 26.8 & 25.8 & 24.7 & 29.1 & 27.0 \\
LLaVA-Med~\cite{li2023llava_med}-7B & 39.9 & 39.1 & 34.6 & 37.4 & 34.0 & 36.9 \\
HuatuoGPT-Vision~\cite{chen2024huatuogpt}-8B &\textcolor{red}{61.0} & 58.8 & \textcolor{red}{50.0} & 44.7 & 38.7 & 49.1 \\
\midrule
\multicolumn{7}{c}{\textit{Finetuned VLM}} \\
Qwen2.5-VL~\cite{bai2025qwen2}-7B (SFT) & 46.4 & 46.7 & 40.0 & 55.6 & 50.0 & 51.7 \\
Qwen2.5-VL~\cite{bai2025qwen2}-7B (GRPO) & \textcolor{blue}{57.1} & \textcolor{red}{66.7} & 30.0 & \textcolor{blue}{70.4} & \textcolor{blue}{63.3} & \textcolor{blue}{57.2} \\
MedCCO-7B & 53.6 & \textcolor{blue}{63.3} & 40.0 & \textcolor{red}{74.1} & \textcolor{red}{66.7} & \textcolor{red}{\textbf{59.3}} \\
\bottomrule
\end{tabular}}
\caption{Generalization (zero-shot) comparison on the selected MMMU~\cite{yue2024mmmu} Health \& Medicine track with category-wise and overall accuracy. The best and second-best results in each column are highlighted in \textcolor{red}{red} and \textcolor{blue}{blue}, respectively. BMS: Basic Medical Science, CM: Clinical Medicine, DLM: Diagnostics and Laboratory Medicine, P: Pharmacy, PH: Public Health.}
\label{tab:mmmu_results}
\end{table*}
\begin{figure}
    \centering
    \includegraphics[width=1\linewidth]{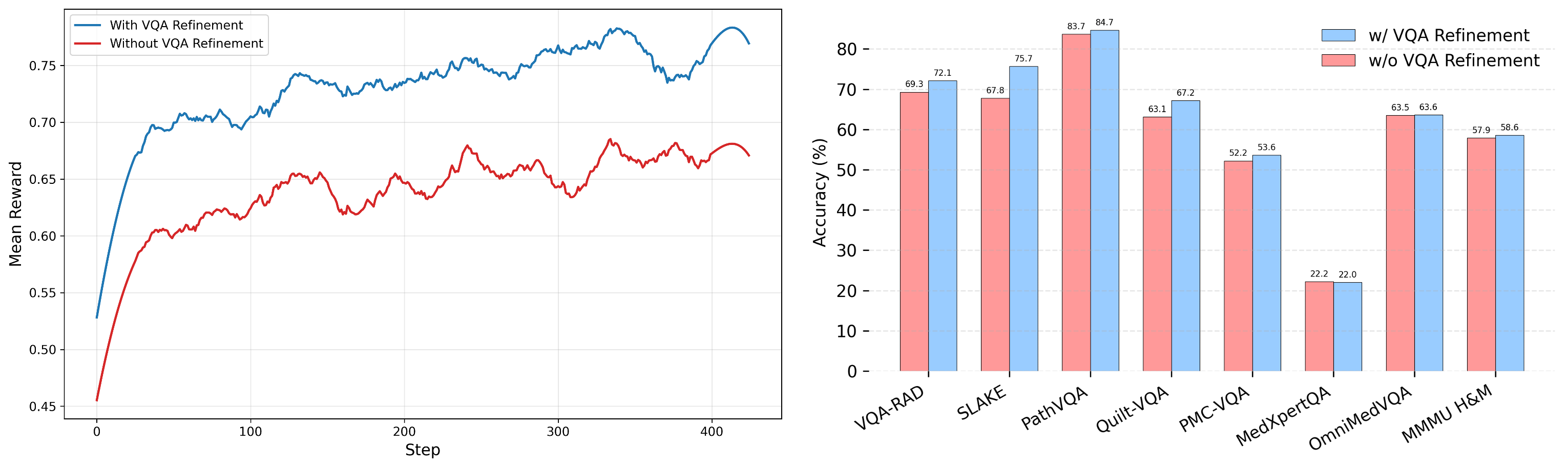}
    \caption{Training curves and overall performance with and without VQA refinement. Incorporating VQA refinement leads to more stable training and improved generalization.}
    \label{fig:qa_refine}
\end{figure}

\begin{figure}[t]
    \centering
    \includegraphics[width=1\linewidth]{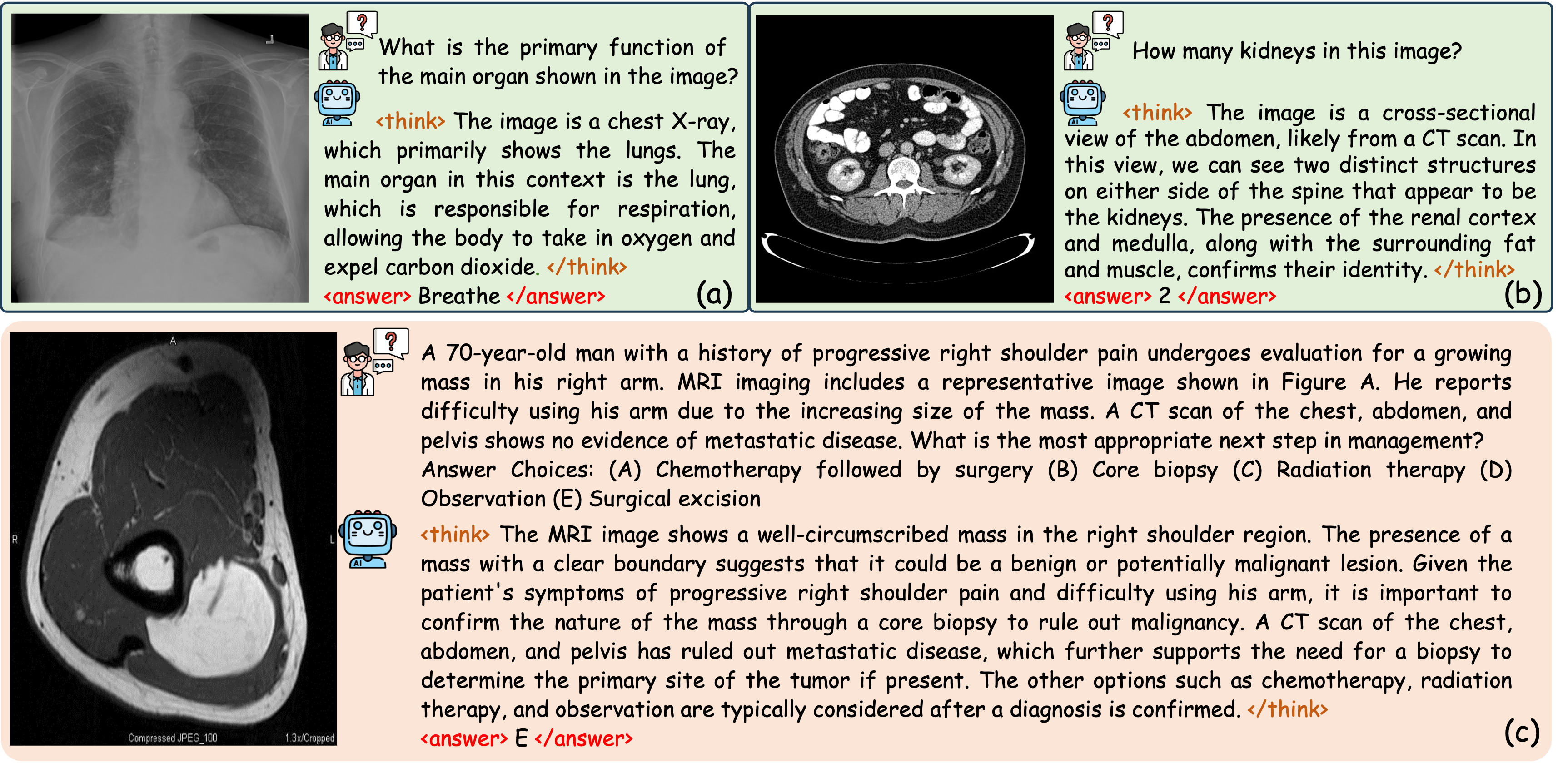}
    \caption{{Qualitative results of open-ended and close-ended VQA reasoning. (a) and (b) show open-ended VQA without an option list, while (c) illustrates a close-ended VQA with an option list. 
    }
}
    \label{fig:case_study}
\end{figure}

\subsection{Ablation Study}
To comprehensively analyze MedCCO's important components, we perform an ablation study on model size, VQA refinement, and fine-tuning type. As for a short demonstration, we report the average performance of OmniMedVQA~\cite{hu2024omnimedvqa}, MedXpertQA~\cite{zuo2025medxpertqa}, and MMMU~\cite{yue2024mmmu} Health\& Medicine. \par

\noindent{\textbf{Model size.}} To investigate the impact of model scale on MedCCO, we compared two parameterizations (3B and 7B). As shown in Table~\ref{tab:ablation}, the 7B model consistently outperforms its 3B counterpart under identical settings, demonstrating superior capacity. Moreover, Curriculum GRPO yields uniform gains on the in-domain test sets (3.9\% for 3B, 2.0\% for 7B) and out-of-domain test sets (2.9\% for 3B, 5.2\% for 7B), underscoring its robustness.
\par

\noindent{\textbf{VQA refinement.}} To assess the impact of VQA consistency refinement for open-ended VQA, we conduct joint GRPO experiments using both the close-ended and w/wo refined open-ended data. Figure~\ref{fig:qa_refine} and Table~\ref{tab:ablation} show that the 3B model improves by 2.5\% on in-domain and 3.1\% on out-of-domain test sets, while the 7B model achieves gains of 3.9\% and 1.2\%, respectively. These results confirm the effectiveness of our VQA-consistency optimization.
\par
\noindent{\textbf{Fine-tuning Type.}} As shown in Table~\ref{tab:ablation}, GRPO-based fine-tuning consistently outperforms SFT across out-of-domain test sets, demonstrating superior generalization capability. While SFT performs slightly better on in-domain evaluations, we attribute this to its strength in learning input-output mappings within the training distribution. In contrast, GRPO encourages reasoning and pathway exploration over direct mapping, making it more robust for unfamiliar or out-of-distribution scenarios.

\noindent{\textbf{Curriculum generalizes better than joint way.}} Table~\ref{tab:ablation} shows consistent performance gains on both in-domain and out-of-domain test sets across 3B and 7B models. We attribute this improvement to the mitigation of reward conflict between close-ended and open-ended tasks. Specifically, close-ended tasks adopt discrete rewards, where small differences in output can lead to large variations in reward signals, resulting in sharper gradient updates. In contrast, open-ended tasks use continuous rewards, which produce smoother and less sensitive gradients. When trained jointly, this discrepancy causes gradient imbalance that hinders stable learning. Curriculum learning alleviates this issue by first training the model on close-ended tasks to build a robust reasoning foundation, then gradually adapting it to open-ended tasks. This staged approach bypasses the gradient conflict and facilitates stable acquisition of open-ended reasoning skills. Additionally, KL divergence regularization penalizes policy deviation, further promoting steady and consistent policy improvement.

\begin{table}[t]
\centering
\resizebox{\textwidth}{!}{
\begin{tabular}{c|c|c|cccc|cccccc}
\hline
\multirow{2}{*}{Model Size} & \multirow{2}{*}{VQA refine} & \multirow{2}{*}{FT type} & \multicolumn{4}{c|}{In-domain test} & \multicolumn{6}{c}{Out-of-domain test} \\
\cline{4-13}
& & & VQA-RAD & SLAKE & PathVQA & Avg & Quilt. & PMC. & MedXpert. & OmniMed. & H\&M & Avg \\
\hline
\multirow{5}{*}{3B}
& -   & -        & 62.2 & 68.3 & 62.4 & 64.3 & 53.4&48.4 & 20.3 & 61.6& 53.8 & 47.5 \\
& \checkmark   & SFT      & 71.7 & 81.7 & 87.5 & 80.3 & 51.6 &49.9 & 21.6 & 61.7 & 51.7 & 47.3 \\
& \checkmark   & GRPO     & 66.5 & 73.6 & 81.1 & 73.7 & 63.5 &47.0 & 22.0 & 61.7 & 51.0 & 49.0 \\
& \XSolidBrush & J.GRPO   & 64.5 & 72.4 & 83.4 & 73.4 & 62.7 &49.1 & 20.6 & 59.3 & 49.7 & 48.3 \\
& \checkmark   & J.GRPO   & 66.5 & 79.3 & 81.9 & 75.9 &65.9 &53.0 & 20.8 & 63.0 & 54.5 & 51.4 \\
& \checkmark   & C.GRPO   & 69.7 
 & 80.5 & 82.5 & 77.6 &62.9 &53.1 & 22.7 & 64.6 & 56.6 & 51.9 \\
\hline
\multirow{5}{*}{7B}
& -   & -        & 67.3 & 71.6 & 65.5 & 68.1 & 54.8&50.4 & 21.9 & 63.5 & 51.7 & 48.5 \\
& \checkmark   & SFT      & 71.3 & 78.6 & 87.8 & 79.2 & 60.9 & 49.2 & 20.3 & 55.7 & 51.7 & 47.6 \\
& \checkmark   & GRPO     & 70.5 & 79.3 & 82.8 & 77.5 & 50.2 & 51.2 & 21.4 & 65.1 & 57.2 & 49.0 \\
& \XSolidBrush  & J.GRPO   & 69.3 &67.8   &83.7   & 73.6   &  63.1   & 52.2    & 22.2    & 63.5    & 57.9  & 51.8 \\
& \checkmark & J.GRPO   & 72.1&75.7   &84.7   & 77.5   &   67.2  & 53.6    & 22.0    & 63.6    & 58.6  & 53.0 \\
& \checkmark   & C.GRPO   & 76.3 & 79.4 & 82.8 & 79.5 & 69.4&53.2 & 23.4 & 65.8 & 59.3 & 54.2 \\
\hline
\end{tabular}}
\caption{Ablation study on model size, VQA refinement, and fine-tuning type. SFT: supervised fine-tuning; J.\@: joint; C.\@: curriculum; Quilt.\@: Quilt-VQA~\cite{seyfioglu2024quilt}; PMC.\@: PMC-VQA~\cite{zhang2023pmc}; MedXpert.\@: MedXpertQA~\cite{zuo2025medxpertqa}; OmniMed.\@: OmniMedVQA~\cite{hu2024omnimedvqa}; H\&M: Health and Medical track of MMMU~\cite{yue2024mmmu}.}
\label{tab:ablation}
\end{table}


\section{Conclusion}
\label{sec:conclustion}
In this paper, we introduce MedCCO, the first multimodal medical reasoning model capable of jointly addressing close-ended and open-ended VQA tasks within a unified framework. By adopting a curriculum training strategy that transitions from structured to open-ended supervision, MedCCO enhances reasoning performance while maintaining training stability, achieving state-of-the-art results across diverse medical benchmarks. More broadly, this work sets a foundation for developing clinically adaptable AI systems that support flexible, context-aware reasoning, moving beyond rigid answer formats toward more nuanced and human-aligned medical understanding.

This study represents an initial attempt to integrate both close-ended and open-ended medical VQA data within a GRPO training framework to enhance the reasoning capabilities of medical VLMs. Our experiments reveal that reinforcement learning with open-ended data presents greater challenges compared to training on close-ended questions with predefined answer options. In particular, enabling VLMs to perform accurate reasoning remains difficult due to limitations in visual perception, domain-specific knowledge, and the tendency to generate hallucinations. Moreover, automatically evaluating the quality of reasoning remains an open problem. These challenges highlight promising directions for future work.


\bibliographystyle{plain}
\bibliography{ref}       

\newpage
\appendix
\begin{center}
    {\large \bf Improving Medical Reasoning with Curriculum-Aware Reinforcement Learning} \\[1em]
    {\large Supplementary Material}
\end{center}
\vspace{1em}


\section{Performance Comparison on OmniMedVQA~\cite{hu2024omnimedvqa} dataset}
\begin{table*}[hb]
\small
\centering
\begin{tabularx}{\textwidth}{l|cccccccc|c}
\toprule
\textbf{Model} & CT & FP & MRI & OCT & Der & Mic & X-Ray & US & \makecell{Avg.} \\
\midrule
\multicolumn{10}{c}{\textit{General VLM}} \\
Yi-VL~\cite{young2024yi}-34B & 39.8 & 57.2 & 51.4 & 70.5 & 54.5 & 61.4 & 64.2 & 40.5 & 54.9 \\
LLaVA-v1.6~\cite{liu2024llavanext}-7B & 40.1 & 39.5 & 54.8 & 58.4 & 54.0 & 48.8 & 53.3 & 47.9 & 49.6 \\
LLaVA-v1.6~\cite{liu2024llavanext}-13B & 40.0 & 43.6 & 47.4 & 63.2 & 58.0 & 50.5 & 59.6 & 42.6 & 50.6 \\
LLaVA-v1.6~\cite{liu2024llavanext}-34B & 50.6 & 63.4 & 60.9 & 68.4 & 65.7 & 62.8 & 74.7 & 44.5 & 61.4 \\
LLaVA-v1.5-LLaMA3-8B & 33.0 & 49.7 & 53.8 & \textcolor{blue}{76.0} & 63.1 & 48.4 & 56.6 & 31.2 & 48.8 \\
Qwen2.5-VL~\cite{bai2025qwen2}-7B & 68.9 & 78.6 & 56.3 &64.4  & 66.5 & \textcolor{red}{68.8} &75.6 & 29.1 & 63.5 \\
\midrule
\multicolumn{10}{c}{\textit{Medical VLM}} \\
Med-Flamingo~\cite{moor2023med} & 34.6 & 33.3 & 27.5 & 26.0 & 28.3 & 28.1 & 30.1 & 33.2 & 30.2 \\
RadFM~\cite{wu2023towards} & 33.3 & 35.0 & 22.0 & 31.3 & 36.3 & 28.0 & 31.5 & 26.1 & 30.5 \\
LLaVA-Med~\cite{li2023llava_med}-7B & 25.3 & 48.4 & 35.9 & 42.1 & 45.2 & 44.0 & 31.7 & \textcolor{blue}{83.7} & 44.5 \\
HuatuoGPT-Vision~\cite{chen2024huatuogpt}-8B & 61.6 & \textcolor{red}{80.2} & \textcolor{red}{65.1} & \textcolor{red}{86.3} & \textcolor{red}{71.6} & 67.4 &\textcolor{red}{81.4} & \textcolor{red}{87.4} & \textcolor{red}{75.1} \\
\midrule
\multicolumn{10}{c}{\textit{Finetuned VLM}} \\
Qwen2.5-VL~\cite{bai2025qwen2}-7B (SFT) & 48.5 & 57.8 & 56.0 & 67.7 & 60.4 & 52.1 & 66.9 & 36.2 & 55.7 \\
Qwen2.5-VL~\cite{bai2025qwen2}-7B (GRPO) & \textcolor{blue}{69.9} & \textcolor{blue}{79.4} & 58.2 & 70.0 & \textcolor{blue}{70.1} & \textcolor{blue}{67.7} & \textcolor{blue}{75.9} & 29.7 & 65.1 \\
MedCCO-7B& \textcolor{red}{71.4} & 73.5 & \textcolor{blue}{62.5} & 63.2 & 69.5 & 66.6 & 74.2 & 37.3 & \textcolor{blue}{65.8} \\
\bottomrule
\end{tabularx}
\caption{Generalization (zero-shot) comparison on the OmniMedVQA~\cite{hu2024omnimedvqa} benchmark across different imaging modalities. Results are reported at the modality level of accuracy. Abbreviations: CT: Computed Tomography; FP: Fundus Photography; MRI: Magnetic Resonance Imaging; OCT: Optical Coherence Tomography; Der: Dermoscopy; Mic: Microscopy; US: Ultrasound.}
\label{tab:omnimedvqa}
\end{table*}

We provide additional experiments on OmniMedVQA~\cite{hu2024omnimedvqa} dataset. As illustrated in Table~\ref{tab:omnimedvqa},
For OmniMedVQA~\cite{hu2024omnimedvqa}, our MedCCO attains a competitive second-best result of 65.8\% (averaged accuracy), surpassing all other models except HuatuoGPT-Vision~\cite{chen2024huatuogpt}-8B(75.1\%). The surprisingly high results of HuatuoGPT-Vision~\cite{chen2024huatuogpt}-8B were achieved by leveraging over 1 million curated medical images, which comprise almost all the public datasets as described in~\cite{chen2024huatuogpt} (largely overlapping with OmniMedVQA~\cite{hu2024omnimedvqa}), while MedCCO is trained with only around 27k medical question-answer pairs (3\% of the counterpart and no overlapping with OmniMedVQA~\cite{hu2024omnimedvqa}), highlighting its remarkable data efficiency and consistent performance across diverse modalities. Also, experimental results show that GRPO-based training method offers limited gains for tasks focused purely on visual perception.

\section{Details of Datasets}
For our experiments, we employ eight open-source medical VQA datasets. The details are provided below:

\noindent{\textbf{VQA-RAD}~\cite{lau2018dataset}} is a radiology-focused medical VQA dataset consisting of 3,064 question-answer (QA) pairs for training and 451 for testing. Among the training samples, there are 1,242 open-ended and 1,823 close-ended questions. The test set contains 272 close-ended and 179 open-ended questions.

\noindent{\textbf{SLAKE}~\cite{liu2021slake}} is a bilingual English-Chinese dataset; we use only the English portion in our experiments. It includes data from three imaging modalities: X-ray, CT, and MRI. After removing Chinese entries, the dataset comprises 4,919 training samples (2,976 open-ended, 1,943 close-ended), 1,053 validation samples (657 open-ended, 422 close-ended), and 1,061 test samples (671 open-ended, 416 close-ended).

\noindent{\textbf{PathVQA}~\cite{he2020pathvqa}} is designed for pathology-related visual question answering, including pathological images, cellular microscopy, and some natural disease-related images. The dataset contains 19,654 training samples (9,553 open-ended, 10,621 close-ended), 6,259 validation samples (2,967 open-ended, 3,435 close-ended), and 6,719 test samples (3,201 open-ended, 3,659 close-ended).

\noindent{\textbf{Quilt-VQA}~\cite{seyfioglu2024quilt}} is a benchmark for pathology visual question answering (VQA), specifically focusing on pathology images. It comprises 344 close-ended and 957 open-ended VQA pairs for evaluation.

\noindent{\textbf{PMC-VQA}~\cite{zhang2023pmc}} contains approximately 227K VQA pairs and 149K medical images, covering a broad spectrum of diseases and radiological modalities. For evaluation, we adopt a subset consisting of 2,000 test examples selected from the validation split of HuatuoGPT-Vision~\cite{chen2024huatuogpt}.

\noindent{\textbf{MedXpertQA}~\cite{zuo2025medxpertqa}} is a recently introduced benchmark designed to assess the complex medical reasoning capabilities of large language models (LLMs) and vision-language models (VLMs) using expert-level questions. It comprises two subsets: a pure-text QA set and a multimodal (MM) VQA set. In our experiments, we use only the MM subset, which includes 2,000 multiple-choice questions.

\noindent{\textbf{MMMU Health \& Medicine Track}~\cite{yue2024mmmu}} is part of the broader MMMU benchmark and is aimed at evaluating the performance of multimodal models across a wide range of medical tasks. It covers multiple subfields, including Basic Medical Science, Clinical Medicine, Diagnostics and Laboratory Medicine, Pharmacy, and Public Health. The test set includes 150 multiple-choice questions, as used in~\cite{chen2024huatuogpt}.

\noindent{\textbf{OmniMedVQA}~\cite{hu2024omnimedvqa}} is constructed from 41 classical medical imaging tasks, reformulated for multimodal evaluation. It is designed to comprehensively assess the visual understanding capabilities of multimodal models in medical imaging. We use the test split adopted in~\cite{chen2024huatuogpt}, which contains 11,124 examples.

\section{VQA consistency refinement} 

\begin{table}[hb]
\centering
\begin{tabular}{p{3.5cm} p{10cm}}
\toprule
\textbf{Field} & \textbf{Content} \\
\midrule
\multicolumn{2}{l}{\textbf{Example 1}} \\
Original Question & What is the main organ in the image? \\
Revised Question & Identify the main organs visible in the image. \\
Answer & Liver, Heart, Spleen, Lung \\
Notes & Clarifies the need to identify multiple organs. \\
\midrule
\multicolumn{2}{l}{\textbf{Example 2}} \\
Original Question & How was this image taken? \\
Revised Question & Identify the imaging modality used to capture this image. \\
Answer & X-Ray \\
Notes & Clarifies the specific imaging technique. \\
\midrule
\multicolumn{2}{l}{\textbf{Example 3}} \\
Original Question & What type of imaging is this? \\
Revised Question & Identify the imaging modality and sequence type shown in the image. \\
Answer & MRI, Diffusion Weighted \\
Notes & Clarifies both modality and sequence type. \\
\bottomrule
\end{tabular}
\caption{Examples of Refined VQA Questions and Answers}
\label{tab:vqa_refinement_vertical}
\end{table}

\noindent{\textbf{Prompt.}} Before training on the open-ended data, we employ Qwen2.5-VL~\cite{bai2025qwen2}-72B as an auditor to refine the consistency of our VQA data. The objective of this refinement step is to guide the model in understanding what to answer, rather than merely learning question-answer mappings during supervised fine-tuning (SFT). We apply this consistency refinement process to the VQA-RAD~\cite{lau2018dataset}, SLAKE~\cite{liu2021slake} (English portion), and PathVQA~\cite{he2020pathvqa} datasets. Figure~\ref{fig:prompt_refine} illustrates the detailed prompt used in this process.

\noindent{\textbf{{VQA refinement examples.}}} Table~\ref{tab:vqa_refinement_vertical} presents three examples illustrating our VQA consistency refinement process. After refinement, the alignment between the question and the answer is significantly improved, which helps the model focus on understanding what to answer rather than merely learning superficial mappings. For additional examples of VQA refinement, please refer to our code.

\begin{figure}[htbp]
\centering
\begin{tcolorbox}[
  title=VQA-Consistency Auditor Prompt,
  colback=gray!5!white,
  colframe=black!75!black,
  width=0.99\textwidth,
  coltitle=white,
  boxrule=0.5pt,
  fonttitle=\bfseries,
  fontupper=\small,
  top=2pt, bottom=2pt, left=4pt, right=4pt,
  sharp corners
]

\texttt{ori\_q}: \texttt{\{Original Question\}} \\
\texttt{ori\_a}: \texttt{\{Answer\}}

\textbf{Role:} \textbf{QA-Consistency Auditor} – an expert data-curator.\\
Your task is to refine \textbf{open-ended visual-question-answering (VQA)} pairs so that the revised question and answer remain logically and granularly consistent. These are \textbf{open-end VQA pairs}, \emph{not} closed-end: do \textbf{not} embed answer choices in the question.

\textbf{Process:}
\begin{enumerate}[label=\arabic*, itemsep=0.35em, topsep=0.3em]
    \item Read the original question (\texttt{ori\_q}).
    \item Ignore the visual content; focus only on the wording of the question and the expected form of the answer.
    \item Internally simulate an expert’s likely free-form answer (\texttt{Expert\_Guess}).
    \item Compare \texttt{Expert\_Guess} to the original answer (\texttt{ori\_a}) to spot missing components or granularity gaps.
    \item Decide on a status:
    \begin{itemize}[itemsep=0pt, topsep=0pt]
        \item \textbf{consistent} – \texttt{ori\_q} already elicits exactly the information found in \texttt{ori\_a}.
        \item \textbf{needs\_fix} – \texttt{ori\_q} is too broad, ambiguous, or does not explicitly request every element found in \texttt{ori\_a}.
        \item \textbf{drop} – The pair is unusable (contradictory, nonsensical, etc.).
    \end{itemize}
    \item If the status is \texttt{needs\_fix}, craft \texttt{new\_q} that:
    \begin{itemize}[itemsep=0pt, topsep=0pt]
        \item Starts with a precise \textbf{action verb} ("Identify", "Describe", "Explain", …).  
        \item Explicitly requests \textbf{every component} required by \texttt{ori\_a}.  
        \item Maintains an \textbf{open-end} format (no yes/no phrasing, no embedded choices).  
        \item Provides a \textbf{1-to-1 mapping}: each phrase in \texttt{ori\_a} must correspond to a clearly stated element in \texttt{new\_q}.  
        \item Matches the \textbf{granularity} of \texttt{ori\_a} exactly—no more, no less.  
        \item Ensures \texttt{new\_a} presents components in the \textbf{same order} that \texttt{new\_q} requests them.
    \end{itemize}
    \item Adjust \texttt{new\_a} only if wording changes are necessary for brevity or clarity; never change the meaning.
\end{enumerate}

\vspace{0.5em}
\textbf{Key Requirements:}
\begin{itemize}[itemsep=0pt, topsep=0pt]
    \item \textbf{Open-ended:} Questions must allow free-form expert responses; never embed answer choices.  
    \item \textbf{Multi-component precision:} If the answer contains multiple elements, the question must explicitly ask for each.  
    \item \textbf{Action-verb prompts:} Begin revised questions with verbs like “Identify”, “Describe”, “Explain”.  
    \item \textbf{Granularity match:} Question scope must match answer specificity exactly.  
    \item \textbf{Order consistency:} Arrange components in \texttt{new\_a} in the same sequence as requested in \texttt{new\_q}.  
    \item \textbf{Answer conciseness:} Keep \texttt{new\_a} as short as possible while fully capturing the meaning.
\end{itemize}

\vspace{0.5em}
\textbf{Output format:} \\
Return \textbf{one} JSON object—nothing else—using this template:
\begin{verbatim}
{
  "status": "consistent | needs_fix | drop",
  "ori_q": "<string>",
  "ori_a": "<string>",
  "new_q": "<string>",
  "new_a": "<string>",
  "notes": "<less than 15 words rationale>"
}
\end{verbatim}

\end{tcolorbox}
\caption{Prompt for refining open-ended VQA consistency.}
\label{fig:prompt_refine}
\end{figure}

\section{Additional Qualitative Results}
\vspace{3mm}
\begin{table}[hb]
  \begin{minipage}{0.99\textwidth}
\centering  
\vspace{-4mm}
\scalebox{0.88}{
\begin{tabular}{l p{12.5cm} }
\toprule
 \multicolumn{2}{l}{\bf Open-ended VQA: Example 1}  \\
\midrule
&  \includegraphics[height=3.5cm]{image.png} \\
\\
\textbf{Question} & Identify the lobe where an abnormal mass is visible in the provided image.
\\
\\
\textbf{Answer} & <think> The image is a chest X-ray, which typically shows the lungs, heart, and other structures of the thoracic cavity. In this X-ray, there appears to be a shadow or opacity on the right side of the image, which could indicate an abnormality. The location of this shadow suggests it is in the right lung. The left lung does not appear to have a similar shadow. Therefore, the abnormal mass is likely in the right lower lobe.</think> \quad \quad <answer> Right upper lobe </answer>
\\
\bottomrule
\end{tabular}
}
\vspace{3mm}
\caption{Open-ended VQA: Example 1.}
\label{sample 1}  
  \end{minipage}
\end{table}

\begin{table}[hb]
  \begin{minipage}{0.99\textwidth}
\centering  
\vspace{-4mm}
\scalebox{0.88}{
\begin{tabular}{l p{12.5cm} }
\toprule
 \multicolumn{2}{l}{\bf Open-ended VQA: Example 2}  \\
\midrule
&  \includegraphics[height=3.5cm]{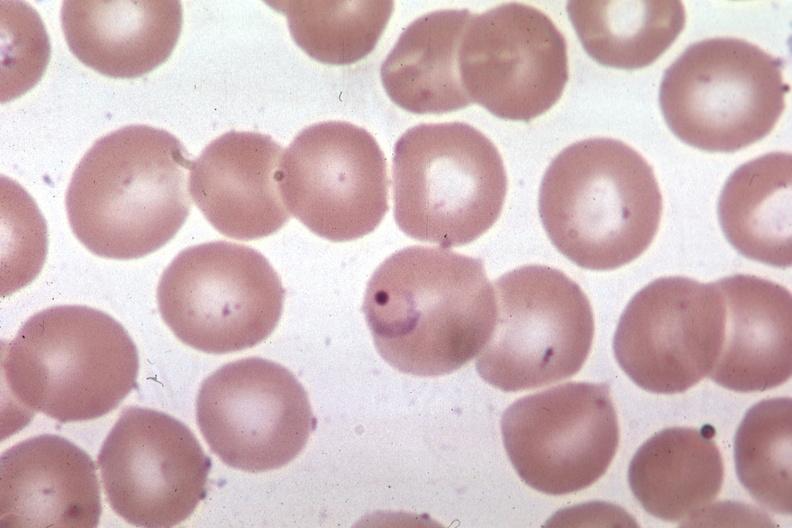} \\

\\
\textbf{Question} & Identify the specific type of malaria parasite present in the image.
\\
\\
\textbf{Answer} & <think> The image shows red blood cells with a single parasite inside one of them, which is characteristic of Plasmodium vivax malaria. The presence of a single parasite within a red blood cell is indicative of the trophozoite stage of the parasite's life cycle. The shape and color of the parasite are consistent with the appearance of Plasmodium vivax. </think>
<answer> Plasmodium vivax </answer>
\\
\bottomrule
\end{tabular}
}
\vspace{1mm}
\caption{Open-ended VQA: Example 2.}
\label{sample 1}  
  \end{minipage}
\end{table}

\begin{table}[ht]
  \begin{minipage}{0.99\textwidth}
\centering  
\vspace{-4mm}
\scalebox{0.88}{
\begin{tabular}{l p{12.5cm} }
\toprule
 \multicolumn{2}{l}{\bf Close-ended VQA: Example 1. }  \\
\midrule
&  \includegraphics[height=3.5cm]{MM-1998-a.jpeg} \\
\\
\textbf{Question} & A 30-year-old male presents to the primary care clinic for evaluation of a painful right-sided scrotal mass. The patient otherwise has no significant past medical history and does not take any daily medications. He denies direct or indirect trauma to the scrotum. He denies tobacco, alcohol, and illicit drug use. Temperature is 98.6 ºF. Blood pressure is 120/80 mmHg, pulse is 80/min, and respiratory rate is 12/min. Palpation of the scrotum demonstrates a unilateral tender mass. Laboratory testing is significant for an elevation in human chorionic gonadotropin and normal alpha-fetoprotein levels. Histopathologic evaluation of the mass is shown in the image below*.* Which of the following is the most accurate regarding the testicular tumor pathology in this patient?
Answer Choices: (A) SALL4 marker negativity (B) Orchiectomy should be avoided (C) Worse prognosis than seminoma (D) Rarely appears as a mixed tumor (E) Analogous to dysgerminoma
\\
\\
\textbf{Answer} & <think> The histopathology shows a pattern consistent with a seminoma, which is a type of testicular cancer. Seminomas are typically characterized by large, round cells with a clear cytoplasm and a prominent nucleolus. The presence of a single type of cell without significant pleomorphism or necrosis is characteristic of seminoma. The elevation in human chorionic gonadotropin (hCG) supports the diagnosis of a germ cell tumor, which includes seminoma. The other options provided do not accurately describe the testicular tumor pathology in this patient based on the given information and the image.</think>\quad \quad <answer> C </answer>
\\
\bottomrule
\end{tabular}
}
\vspace{1mm}
\caption{Close-ended VQA: Example 1.}
\label{sample 1}  
  \end{minipage}
\end{table}

\begin{table}[ht]
  \begin{minipage}{0.99\textwidth}
\centering  
\vspace{-4mm}
\scalebox{0.88}{
\begin{tabular}{l p{12.5cm} }
\toprule
 \multicolumn{2}{l}{\bf Close-ended VQA: Example 2}  \\
\midrule
&  \includegraphics[height=3.5cm]{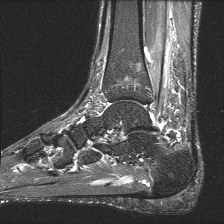} \\
\\
\textbf{Question} & What can be observed in this image?
A. Shin splints B. Tennis elbow C. Plantar fascia pathology D. Meniscal tear
\\
\\
\textbf{Answer} & <think> The image appears to be an MRI scan of the foot, which is commonly used to diagnose soft tissue injuries. The presence of fluid or abnormal signal intensity around the plantar fascia suggests inflammation or pathology in that area. This is characteristic of plantar fasciitis, a condition often referred to as \"plantar fascia pathology.\" There is no indication of shin splints, tennis elbow, or meniscal tear in this image.</think> \quad \quad \quad   \quad  \quad \quad\quad\quad\quad\quad <answer> C </answer>
\\
\bottomrule
\end{tabular}
}
\vspace{1mm}
\caption{Close-ended VQA: Example 2.}
\label{sample 1}  
  \end{minipage}
\end{table}

\end{document}